\documentclass[10pt, a4paper]{article}
\usepackage{amsmath,amssymb}
\usepackage{booktabs}
\usepackage{graphicx}
\usepackage{hyperref}
\usepackage{array}
\usepackage{caption}
\usepackage[utf8]{inputenc}
\usepackage[T1]{fontenc}
\usepackage{subcaption}
\usepackage{xcolor}
\usepackage{multirow}
\usepackage{float}
\usepackage{titling}  

\newcommand{\model}{ZACH-ViT}

\title{\model: Regime-Dependent Inductive Bias in Compact Vision Transformers for Medical Imaging}

\author{%
Athanasios Angelakis\\[6pt]
{\small \href{mailto:athanasios.angelakis@unibw.de}{athanasios.angelakis@unibw.de}}\\[2pt]
{\small ORCID: \href{https://orcid.org/0000-0003-1226-9560}{0000-0003-1226-9560}}\\[6pt]
{\small $^{1}$ BioML Lab, RI CODE, UniBw, Munich, Germany}\\[2pt]
{\small $^{2}$ EDS, Amsterdam UMC, Amsterdam, Netherlands}%
}

\pretitle{\begin{center}\LARGE}
\posttitle{\par\end{center}\vskip 0.5em}
\preauthor{\begin{center}\large}
\postauthor{\par\end{center}}
\predate{\begin{center}}
\postdate{\par\end{center}}

\date{}
\begin{document}
\maketitle
\begin{abstract}
Vision Transformers rely on positional embeddings and class tokens that encode fixed spatial priors. While effective for natural images, these priors may be suboptimal when spatial layout is weakly informative or inconsistent, a frequent condition in medical imaging. We introduce \model\ (Zero-token Adaptive Compact Hierarchical Vision Transformer), a compact Vision Transformer that removes both positional embeddings and the \texttt{[CLS]} token, achieving permutation-invariant patch processing through global average pooling over patch representations. The term ``Zero-token'' refers specifically to the removal of the dedicated \texttt{[CLS]} aggregation token and positional embeddings; patch tokens are still present and processed normally. Adaptive residual projections preserve training stability in compact configurations while maintaining a strict parameter budget.

Evaluation is performed across seven MedMNIST datasets spanning binary and multi-class tasks under a strict few-shot protocol (50 samples per class, fixed hyperparameters, and five random seeds). The empirical analysis reveals a clear regime-dependent behavior: \model\ (0.25M parameters, trained from scratch) achieves its strongest advantage on BloodMNIST and remains competitive on PathMNIST, while its relative advantage decreases on datasets with stronger anatomical priors, such as OCTMNIST and OrganAMNIST, consistent with the architectural hypothesis. Additional component and pooling ablations show that positional support becomes mildly beneficial as spatial structure increases, whereas reintroducing a dedicated \texttt{[CLS]} token is consistently unfavorable. These results support the view that architectural alignment with data structure can be more important than pursuing universal benchmark dominance. Despite its minimal size and lack of pretraining, \model\ achieves competitive performance under data-scarce conditions, making it relevant for compact medical imaging systems and low-resource settings. Code and models are available at \url{https://github.com/Bluesman79/ZACH-ViT}.
\end{abstract}


\section{Introduction}
\label{sec:intro}

Vision Transformers (ViTs) have reshaped computer vision through strong performance and scalability on large-scale benchmarks such as ImageNet~\cite{dosovitskiy2021}. Their architecture typically relies on positional embeddings and a \texttt{[CLS]} token, introducing an inductive bias toward spatially structured representations. While appropriate for natural images, this assumption does not universally hold. In many real-world scenarios, image patches lack intrinsic ordering or spatial relationships carry limited discriminative value.

Several medical imaging modalities illustrate this mismatch. Blood cells appear randomly distributed within microscopy fields, while histopathology patches often behave as unordered collections where diagnosis depends more on cellular composition than on absolute layout. Even in modalities with stronger anatomical organization, acquisition variability and local texture can limit the usefulness of rigid positional assumptions. In such settings, positional priors may encourage models to exploit unstable spatial correlations rather than invariant visual characteristics.

To address this mismatch between architectural priors and data structure, we introduce \model\ (Zero-token Adaptive Compact Hierarchical Vision Transformer), a compact architecture designed around permutation-invariant patch processing. Here, ``Zero-token'' refers specifically to the absence of a dedicated aggregation token and positional priors, not to the removal of patch tokens themselves:
\begin{itemize}
\item \textbf{Zero-token}: positional embeddings are removed and the \texttt{[CLS]} aggregation token is replaced by global average pooling, allowing patch tokens to be treated as an unordered set.
\item \textbf{Adaptive}: residual projections preserve gradient stability when feature dimensionality changes across transformer layers.
\item \textbf{Compact}: the architecture operates with 0.25M parameters by avoiding components dedicated to modeling potentially weakly informative spatial structure.
\item \textbf{Hierarchical}: multiple transformer layers capture compositional features without introducing positional bias.
\item \textbf{End-to-end}: unlike MIL-based transformers designed primarily for bag-level aggregation, \model\ functions as a standalone compact vision backbone with direct patch-level representation learning.
\end{itemize}

The resulting model is permutation-invariant by construction while remaining parameter-efficient and computationally lightweight. We evaluate \model\ across seven representative MedMNIST datasets~\cite{yang2023medmnistv2}, covering binary classification (BreastMNIST, PneumoniaMNIST) and multi-class tasks (BloodMNIST, DermaMNIST, OCTMNIST, PathMNIST, OrganAMNIST). These datasets span a spectrum of spatial structure strength, ranging from weakly structured blood microscopy images to anatomically constrained abdominal and retinal scans.

Rather than claiming universal superiority, our results reveal a regime-dependent behavior: \model\ performs strongest when spatial order is weakly informative (BloodMNIST, PathMNIST) and remains competitive when anatomical structure imposes more stable spatial relationships (OCTMNIST, OrganAMNIST). This supports the broader hypothesis that architectural inductive biases should align with the spatial characteristics of the target data instead of assuming uniform spatial relevance.

This perspective is relevant for medical imaging settings characterized by memory, compute, and data constraints, where robustness and parameter efficiency can be as important as absolute benchmark performance. In such settings, overly strong architectural priors may be as limiting as insufficient model capacity.

Our contributions can be summarized as follows:
\begin{itemize}
\item We introduce \model, a compact permutation-invariant Vision Transformer that removes positional embeddings and token-based aggregation, enabling efficient end-to-end patch processing without reliance on fixed spatial priors.
\item We provide a systematic regime-spectrum analysis linking transformer inductive bias to spatial structure strength, demonstrating when permutation invariance becomes advantageous under controlled few-shot medical imaging conditions.
\item Through a comprehensive benchmark across fifteen architectures and seven MedMNIST datasets under identical protocols, together with targeted component and pooling ablations, we show that architectural alignment with data structure can be more influential than absolute model scale or pretraining.
\end{itemize}

Overall, the main contribution of this work extends beyond the architecture itself: we establish a reproducible framework for studying when permutation-invariant transformers are appropriate in medical imaging, shifting the focus from absolute benchmark dominance toward principled inductive-bias alignment.

\section{Related Work}
\label{sec:rel_work}

\subsection{Permutation-Invariant Vision Architectures}

A preliminary arXiv preprint introduced the ZACH-ViT architecture on a clinical lung ultrasound dataset~\cite{angelakis2025zachvit}. Subsequent analysis revealed annotation issues in that dataset, preventing reliable clinical conclusions. The present work therefore serves as the first controlled validation of \model\ across seven MedMNIST datasets spanning weak to strong spatial structure under a unified few-shot protocol. This broader evaluation enables a systematic analysis of how permutation-invariant architectural choices behave across regimes with different degrees of spatial organization.

More generally, permutation-invariant processing has appeared in multiple-instance learning (MIL), set-based modeling, and architectures designed for weakly ordered inputs. The contribution of the present work is not to claim novelty for permutation invariance itself, but to study how removing positional priors within a compact end-to-end transformer backbone interacts with spatial-structure regimes in medical imaging.

\subsection{Distinction from MIL-Based Transformers}

TransMIL~\cite{shao2021transmil} also processes unordered sets of patch representations within a multiple-instance learning framework, yielding permutation-invariant aggregation at the bag level. However, its architectural role differs substantially from that of \model. TransMIL is primarily designed as a transformer-based MIL aggregator over pre-extracted instances, particularly for whole-slide imaging. In contrast, \model\ is a compact end-to-end vision backbone that performs direct patch-level representation learning from images without an MIL preprocessing stage.

Thus, although both methods can operate on unordered collections, they differ in objective, architectural role, and deployment assumptions. The contribution of \model\ lies not in introducing permutation invariance per se, but in showing how a minimal transformer architecture without positional priors behaves across datasets with different levels of spatial structure.

\subsection{Positional Embedding Removal Across Domains}

The idea that positional encodings may act as unnecessary or even harmful inductive biases has recently emerged in multiple domains. In language modeling, Gelberg et al.~\cite{gelberg2025drope} showed that removing rotary positional embeddings after pretraining can improve context-length extrapolation in large language models. Although developed in a different setting, this observation is conceptually aligned with our work in questioning whether positional information should always be embedded as a permanent architectural prior.

Despite this conceptual overlap, the underlying challenges differ substantially. DroPE addresses sequence extrapolation in language models, where positional information is intrinsic but becomes out-of-distribution at longer context lengths. In contrast, \model\ targets medical imaging scenarios in which spatial information may be weakly informative or non-diagnostic by design. Taken together, these directions suggest a broader principle: architectural components encoding spatial or temporal priors should be matched to the structural properties of the data rather than applied universally. Our regime-spectrum analysis provides empirical support for this principle within medical vision tasks.

\section{Method: \model}
\label{sec:method}
\subsection{Architecture Overview}
Given an input image $X \in \mathbb{R}^{H \times W \times C}$, \model\ extracts non-overlapping patches and projects them into a latent space:
\[
Z_0 = \text{Linear}(\text{Patchify}(X)) \in \mathbb{R}^{N \times d},
\]
where $N$ denotes the number of patches and $d$ the embedding dimension. Unlike standard ViTs, no positional embeddings are added, preserving invariance to patch permutations. A stack of transformer blocks processes $Z_0$, and the final representation is obtained via global average pooling:
\[
h = \frac{1}{N}\sum_{i = 1}^{N}Z_{L}^{(i)}
\]
A linear classifier maps $h$ to output logits. By removing positional encoding and replacing the \texttt{[CLS]} token with global pooling, permutation invariance is achieved by construction while avoiding parameters dedicated to modeling potentially non-informative spatial relationships.

\subsection{Adaptive Residual Projections}
Compact transformer architectures may exhibit unstable optimization when the feature dimensionality of the residual branch and the transformed branch differs, particularly under strict parameter constraints. Let $x$ denote the input to a transformer block and $y$ its transformed output. Residual summation is defined as
\[
x \leftarrow
\begin{cases}
W_{\text{proj}} x + y, & \text{if } \dim(x) \neq \dim(y), \\
x + y, & \text{otherwise},
\end{cases}
\]
where $W_{\text{proj}}$ is a learnable linear projection initialized to zero, and $\dim(\cdot)$ denotes feature dimensionality. This mechanism preserves gradient flow across dimensional transitions while introducing minimal additional parameters, enabling stable training within the 0.25M-parameter regime. In preliminary experiments without adaptive projections, compact variants with changing feature dimensionality showed occasional optimization instability (training divergence or increased variance across seeds), motivating the inclusion of this lightweight projection mechanism.

\section{Experimental Protocol}
\label{sec:protocol}

\subsection{Few-Shot Setting}

All experiments follow a unified protocol to ensure consistent comparison:
\begin{itemize}
\item 50 training samples per class (randomly sampled from the official training split)
\item Validation and test splits kept unchanged
\item Batch size: 16
\item Learning rate: $1 \times 10^{-4}$ (Adam optimizer)
\item Epochs: 23
\item Random seeds: $\{3,5,7,11,13\}$ (reported as mean $\pm$ standard deviation)
\end{itemize}

This setup intentionally stresses models under data-scarce conditions representative of real-world medical imaging scenarios.

\subsection{Spatial Structure Spectrum and Dataset Selection}

We evaluate seven datasets from the MedMNIST v2 suite~\cite{yang2023medmnistv2}, selected to span a broad spectrum of spatial structure while maintaining task compatibility:
\begin{itemize}
\item \textbf{Very weak (1):} BloodMNIST -- blood cells appear randomly distributed in microscopy fields.
\item \textbf{Weak (2):} PathMNIST -- histopathology patches behave as unordered diagnostic collections.
\item \textbf{Moderate (3):} BreastMNIST (variable ultrasound probe position), PneumoniaMNIST (inconsistent X-ray framing).
\item \textbf{Strong (4-5):} DermaMNIST (lesion-centered but variable), OCTMNIST (fixed retinal layer structure), OrganAMNIST (stable abdominal anatomy).
\end{itemize}

Three datasets from MedMNIST were excluded for methodological reasons:
\begin{itemize}
\item \textbf{ChestMNIST:} multi-label classification task, incompatible with the unified binary/multi-class evaluation protocol.
\item \textbf{RetinaMNIST:} ordinal regression task requiring specialized evaluation metrics.
\item \textbf{TissueMNIST:} histopathology modality with visual characteristics highly overlapping PathMNIST, adding limited additional structural diversity.
\end{itemize}

This selection preserves consistent evaluation metrics while covering the intended spatial-structure spectrum, from weakly ordered microscopy-like settings to more anatomically constrained imaging regimes.

\subsection{Model Families and Benchmarking}
\label{subsec:model_selection}
We evaluate fifteen models spanning four architectural families, covering different parameter scales, initialization regimes, and inductive biases.

Rather than reporting only selected baselines, all trained models are included in the analysis. This enables comparison from ultra-compact scratch-trained architectures (0.09M parameters) to large pretrained Transformers (85.8M parameters).

Table~\ref{tab:all_models} summarizes the evaluated models.

\begin{table}[t]
\centering
\setlength{\tabcolsep}{4pt}
\caption{Complete list of evaluated models trained under the identical few-shot protocol. Initialization indicates whether ImageNet-pretrained weights were used. Parameter counts are exact. Disk footprint reports saved weight file size measured on BreastMNIST (seed=7, 50-shot, batch size 16, 23 epochs) to approximate deployment-oriented storage cost. Citations correspond to original architecture publications.}
\label{tab:all_models}
\resizebox{\textwidth}{!}{
\begin{tabular}{@{}l l c c r@{}}
\toprule
\textbf{Family} & \textbf{Model} & \textbf{Params (M)} & \textbf{Initialization} & \textbf{Weights (MB)} \\
\midrule
\multirow{5}{*}{Scratch Models}
& ABMIL \cite{ilse2018abmil} & 0.09 & Random & 1.082 \\
& \textbf{ZACH-ViT (Ours)} \cite{angelakis2025zachvit} & 0.25 & Random & \textbf{2.949} \\
& TransMIL \cite{shao2021transmil} & 0.26 & Random & 3.051 \\
& Minimal-ViT \cite{dosovitskiy2021} & 0.62 & Random & 7.458 \\
& CNN-ABMIL \cite{ilse2018abmil,he2016resnet} & 1.12 & Random & 103.205 \\
\midrule
\multirow{5}{*}{CNN (ImageNet)}
& MobileNetV2 \cite{sandler2018mobilenetv2} & 2.39 & ImageNet & 27.959 \\
& EfficientNetB0 \cite{tan2019efficientnet} & 4.17 & ImageNet & 48.611 \\
& DenseNet121 \cite{huang2017densenet} & 7.09 & ImageNet & 82.666 \\
& InceptionV3 \cite{szegedy2016inceptionv3} & 22.03 & ImageNet & 253.120 \\
& ResNet50 \cite{he2016resnet} & 23.80 & ImageNet & 273.133 \\
\midrule
\multirow{3}{*}{Transformers (ImageNet)}
& DeiT-Small \cite{touvron2021deit} & 21.67 & ImageNet & 82.722 \\
& Swin-Tiny \cite{liu2021swin} & 27.52 & ImageNet & 105.058 \\
& ViT-B/16 \cite{dosovitskiy2021} & 85.80 & ImageNet & 327.365 \\
\midrule
\multirow{2}{*}{Hybrid / Modern Backbones (ImageNet)}
& MambaOut-Tiny \cite{zhang2024mambaout} & 24.24 & ImageNet & 92.555 \\
& ConvNeXt-Tiny \cite{liu2022convnext} & 27.92 & ImageNet & 319.985 \\
\bottomrule
\end{tabular}
}
\end{table}

Subset-specific comparisons (e.g., scratch-only analyses to isolate inductive bias or reduced plots for readability) are explicitly indicated and do not reflect selective reporting.

Disk footprint complements parameter count by measuring serialized model size under a fixed training configuration, reflecting practical storage and distribution cost relevant to edge deployment and reproducibility.

\subsection{Implementation and Experimental Environment}

All experiments were executed in a controlled Linux environment using Python 3.10.12. ZACH-ViT and all ablation studies were implemented in TensorFlow 2.14.0.

Certain baseline architectures whose official or stable implementations are available only in PyTorch were evaluated using their original implementations (PyTorch 2.2.2, timm 1.0.24). These models were executed under identical dataset splits, training protocols, and hardware settings to ensure comparability.

Experiments were run on a workstation equipped with an NVIDIA RTX A5000 GPU (24\,GB VRAM), CUDA 12.8, and 128 CPU cores. The software stack included library\_zViT (version 2026001011), MedMNIST 3.0.2, NumPy 1.24.3, Pandas 1.5.3, Matplotlib 3.10.8, and Scikit-learn 1.2.0.

Random seeds were fixed across runs, and all models used identical preprocessing and data splits to maximize reproducibility.

\subsection{Evaluation Metrics}

Primary evaluation metrics depend on dataset type:
\begin{itemize}
\item \textbf{Binary datasets} (BreastMNIST, PneumoniaMNIST): Test AUC@0.5 (threshold fixed at 0.5).
\item \textbf{Multi-class datasets}: Test MacroF1 to mitigate class imbalance effects.
\end{itemize}

We additionally report generalization gap (Train $-$ Test) to quantify overfitting and, where relevant, deployment-oriented efficiency indicators such as parameter count, serialized model size, and inference behavior.

\section{Results on Medical Imaging Benchmarks}
\label{sec:medical_experiments}

Unless stated otherwise, all results in this section are reported for the full set of fifteen models summarized in Table~\ref{tab:all_models}.

\subsection{Regime Spectrum Validation: When Permutation Invariance Matters}
\label{subsec:regime_spectrum}

\begin{figure}[t]
\centering
\includegraphics[width=0.95\linewidth]{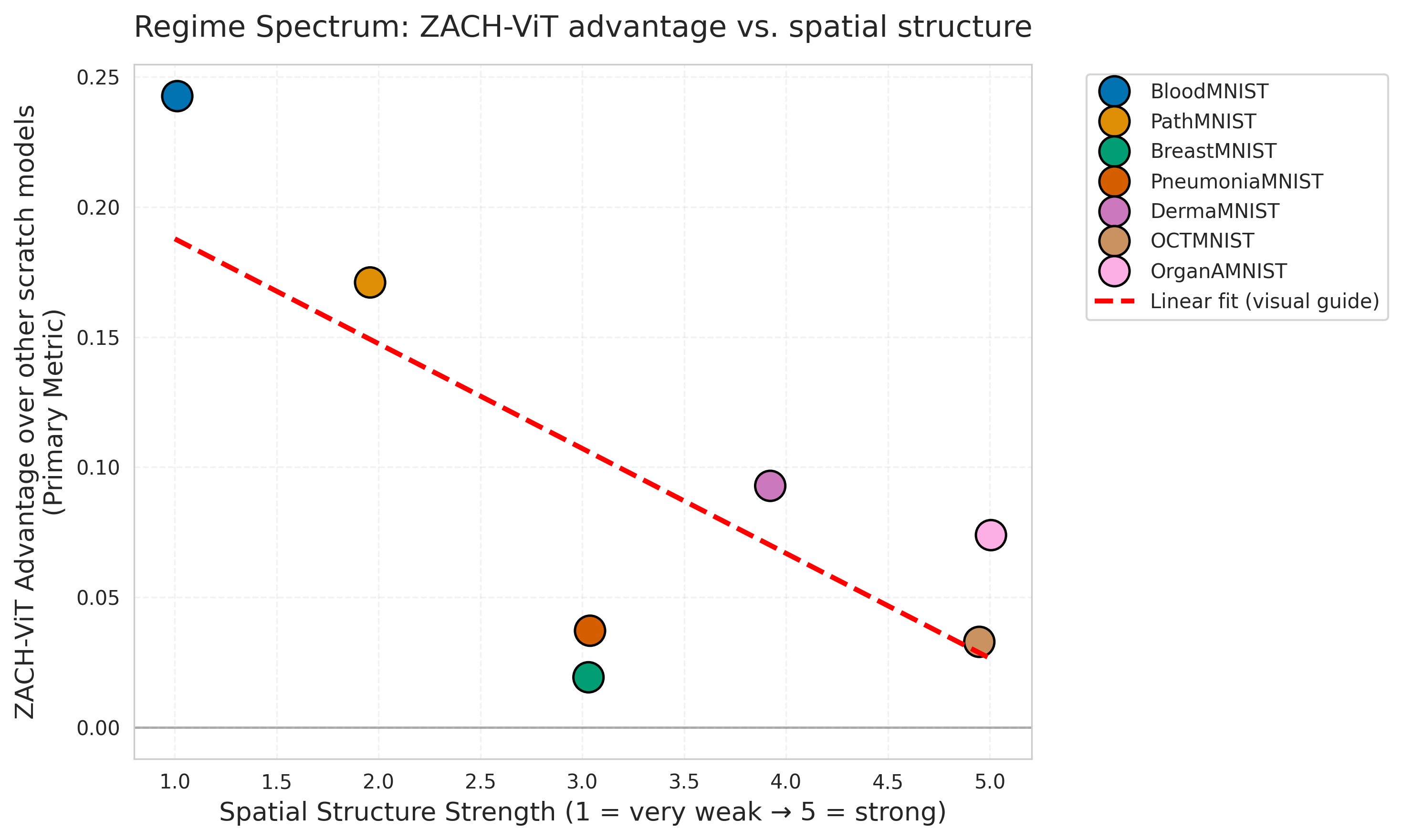}
\caption{Regime spectrum analysis. Each point shows ZACH-ViT's advantage over the mean performance of other scratch-trained baselines on a dataset. Datasets are ordered by an ordinal spatial-structure strength index (1 = very weak, 5 = strong). The trend suggests that ZACH-ViT's relative advantage is larger when spatial layout is weakly informative and smaller when anatomical structure is fixed. The downward trend confirms the hypothesis that ZACH-ViT's permutation-invariant bias is most beneficial when spatial layout is weakly informative.} 
\label{fig:regime_spectrum}
\end{figure}

Our architectural hypothesis predicts that permutation invariance is most beneficial when spatial layout is weakly informative. Figure~\ref{fig:regime_spectrum} evaluates this relationship across seven MedMNIST datasets ordered by spatial structure strength.

\model\ achieves the strongest performance among sub-1M models on BloodMNIST (MacroF1 $0.600 \pm 0.071$) and PathMNIST ($0.578 \pm 0.037$), datasets where spatial arrangement is weakly informative. In these settings, permutation-invariant processing aligns with the data characteristics, yielding a +0.051 MacroF1 gain over TransMIL on BloodMNIST, while achieving comparable (slightly lower) performance on PathMNIST (-0.019 MacroF1).

Performance differences decrease on OCTMNIST and OrganAMNIST, where anatomical structure imposes strong spatial organization ($0.247 \pm 0.052$ and $0.436 \pm 0.036$ MacroF1). This trend supports the hypothesis: positional information contributes positively when spatial relationships are diagnostically relevant. In such cases, removing positional priors may limit modeling of fine-grained structural patterns (e.g., retinal layer organization). Overall, these results clarify the regimes in which permutation invariance is advantageous.

\subsection{Parameter Efficiency Without Pretraining}
\label{subsec:param_efficiency}

\begin{figure}[t]
\centering
\includegraphics[width=0.95\linewidth]{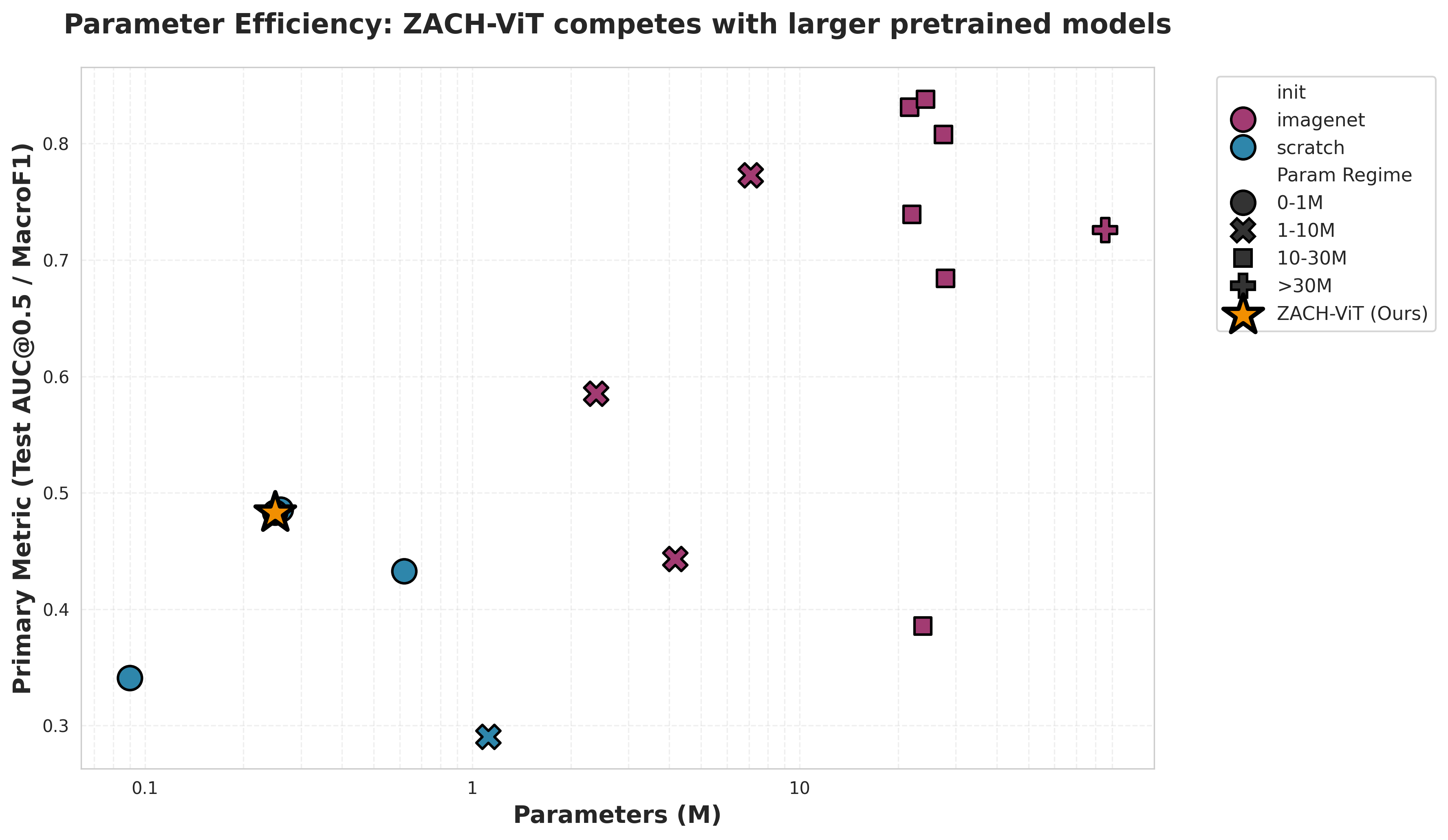}
\caption{Global parameter efficiency across all evaluated models. \model\ competes with substantially larger pretrained architectures despite scratch training.}
\label{fig:param_efficiency}
\end{figure}

Medical edge deployment often constrains model size and limits reliance on large pretrained networks. Figure~\ref{fig:param_efficiency} shows that \model\ achieves competitive performance with only 0.25M parameters. On BreastMNIST, performance is comparable to MobileNetV2 (2.39M parameters) despite a substantially smaller parameter budget and no pretraining. On PathMNIST, \model\ outperforms all sub-10M models except DenseNet121.

\begin{figure}[t]
\centering
\includegraphics[width=\linewidth]{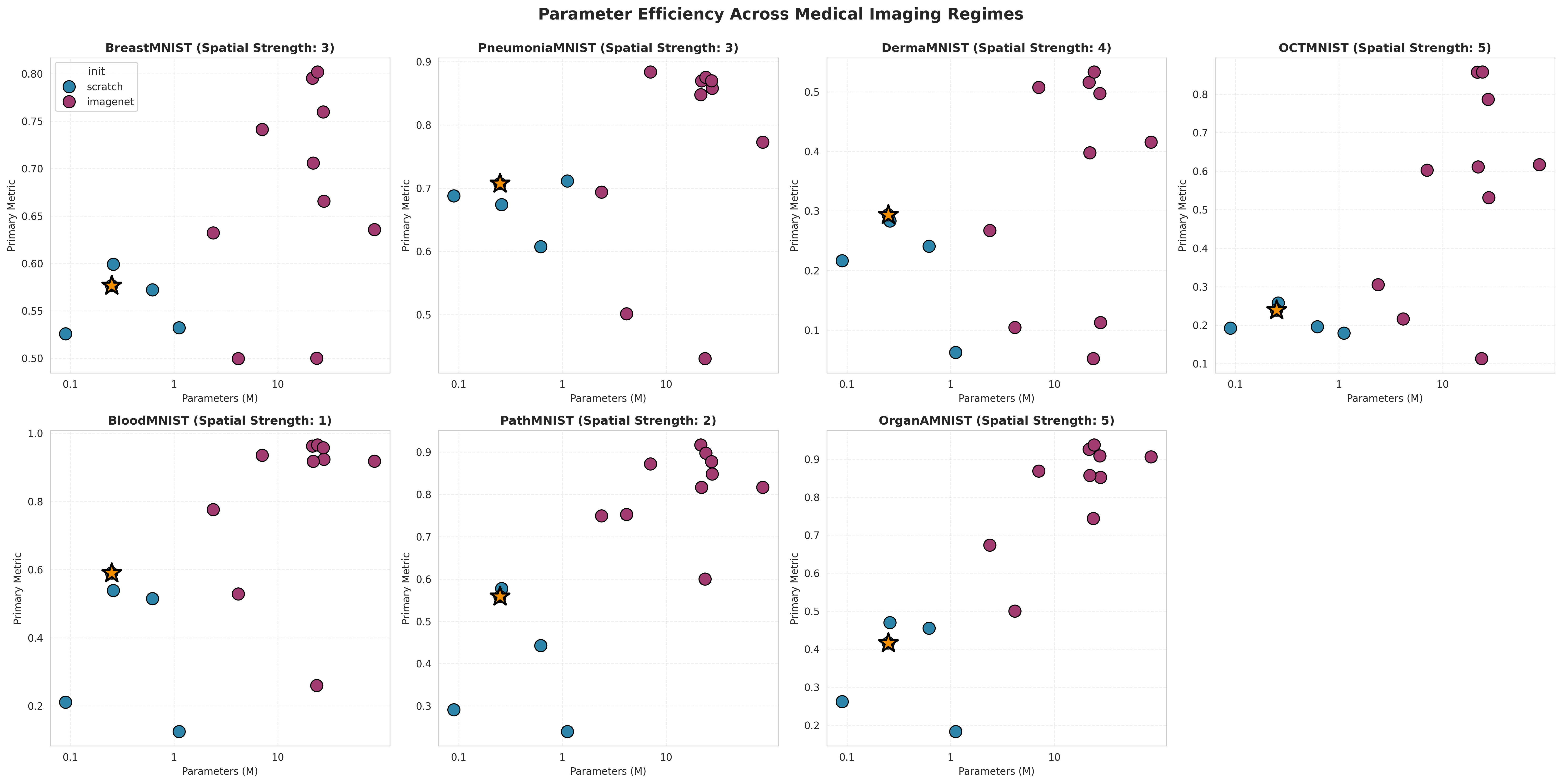}
\caption{Parameter efficiency across individual MedMNIST datasets. Each subplot reports model performance versus parameter count within a specific spatial-structure regime. The star indicates \model. Results illustrate regime-dependent efficiency: \model\ shows strong competitiveness in weak-structure datasets, while pretrained models generally achieve higher performance in strongly structured anatomical regimes.}
\label{fig:param_efficiency_dataset}
\end{figure}

To better understand how parameter efficiency varies across spatial-structure regimes, Figure~\ref{fig:param_efficiency_dataset} provides a per-dataset decomposition. Each subplot reports model performance versus parameter count within a specific MedMNIST dataset, illustrating how the efficiency-performance trade-off changes depending on spatial structure strength. The star marker highlights \model. The visualization shows that \model\ remains consistently competitive among scratch-trained models across datasets, with strongest relative efficiency in weak-structure regimes (BloodMNIST, PathMNIST), while pretrained models tend to dominate regimes with stronger anatomical priors (OCTMNIST, OrganAMNIST).

This decomposition reveals that parameter efficiency is not a global property but a regime-dependent one: the same compact architecture occupies different positions on the efficiency-performance frontier depending on spatial structure strength. Consequently, evaluating edge-oriented architectures solely through aggregated performance can obscure where architectural inductive biases are most beneficial.

Importantly, although both \model\ and TransMIL exhibit permutation-invariant behavior, their architectural roles differ. TransMIL is primarily designed as a transformer-based multiple-instance learning aggregation framework operating at the bag level, whereas \model\ functions as a compact end-to-end vision backbone with direct patch-level feature learning and adaptive residual projections. This distinction helps explain why both models achieve comparable global ranking while exhibiting different behavior across spatial regimes and parameter-efficiency trade-offs.

This efficiency follows from removing explicit spatial encoding components and allocating capacity toward feature representation rather than positional modeling. Table~\ref{tab:summary_stats} summarizes performance across all seven datasets, showing strong results in weak-structure regimes and competitive performance elsewhere.

\begin{table}[t]
\centering
\caption{Test MacroF1/AUC@0.5 (mean $\pm$ std) for scratch models across MedMNIST7. Bold indicates best performer per dataset.}
\label{tab:summary_stats}
\resizebox{\textwidth}{!}{
\begin{tabular}{@{}lcccccc@{}}
\toprule
Dataset & Spatial Strength & \model & ABMIL & CNN-ABMIL & Minimal-ViT & TransMIL \\
\midrule
BloodMNIST & 1 (weakest) & \textbf{0.600 $\pm$ 0.071} & 0.211 $\pm$ 0.050 & 0.125 $\pm$ 0.067 & 0.515 $\pm$ 0.086 & 0.538 $\pm$ 0.065 \\
PathMNIST & 2 & \textbf{0.578 $\pm$ 0.041} & 0.291 $\pm$ 0.084 & 0.239 $\pm$ 0.028 & 0.443 $\pm$ 0.061 & 0.577 $\pm$ 0.048 \\
BreastMNIST & 3 & \textbf{0.577 $\pm$ 0.018} & 0.526 $\pm$ 0.058 & 0.532 $\pm$ 0.072 & 0.572 $\pm$ 0.014 & 0.599 $\pm$ 0.027 \\
PneumoniaMNIST & 3 & \textbf{0.707 $\pm$ 0.017} & 0.688 $\pm$ 0.026 & 0.711 $\pm$ 0.042 & 0.607 $\pm$ 0.027 & 0.674 $\pm$ 0.019 \\
DermaMNIST & 4 & \textbf{0.293 $\pm$ 0.029} & 0.216 $\pm$ 0.023 & 0.062 $\pm$ 0.058 & 0.241 $\pm$ 0.033 & 0.283 $\pm$ 0.024 \\
OCTMNIST & 5 (strongest) & 0.239 $\pm$ 0.056 & 0.192 $\pm$ 0.059 & 0.179 $\pm$ 0.016 & 0.196 $\pm$ 0.069 & \textbf{0.258 $\pm$ 0.053} \\
OrganAMNIST & 5 & 0.416 $\pm$ 0.037 & 0.262 $\pm$ 0.023 & 0.183 $\pm$ 0.055 & \textbf{0.455 $\pm$ 0.029} & 0.470 $\pm$ 0.057 \\
\bottomrule
\end{tabular}
}
\end{table}

\subsection{Generalization and Efficiency}
\label{subsec:edge_deployment}

\begin{figure}[t]
\centering
\includegraphics[width=0.95\linewidth]{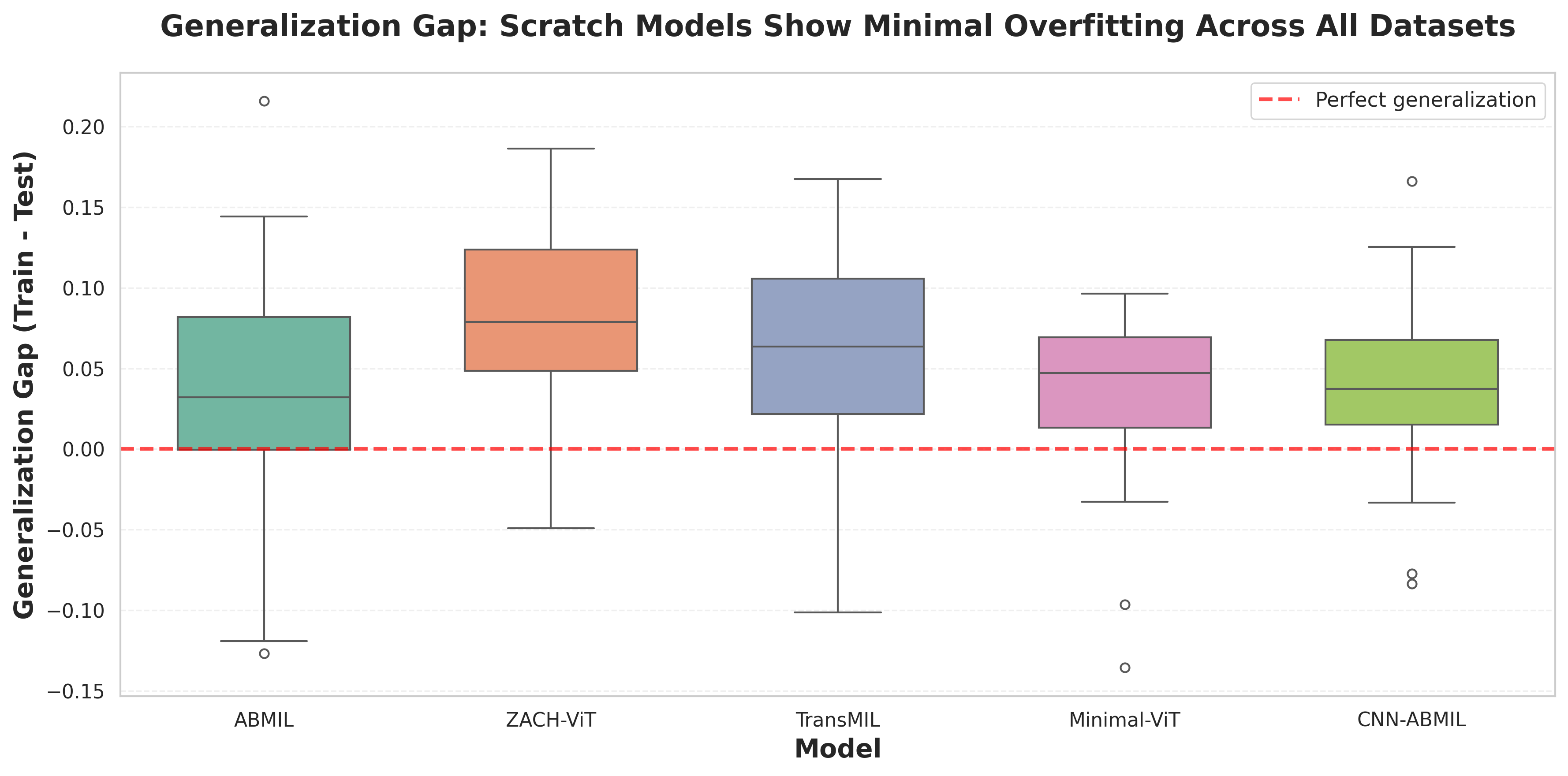}
\caption{Generalization gap (Train-Test) for scratch models. \model\ exhibits consistently small gaps across datasets.}
\label{fig:overfitting_analysis}
\end{figure}

A practically useful compact architecture should exhibit both stable generalization and favorable efficiency characteristics. Figure~\ref{fig:overfitting_analysis} shows that \model\ maintains small train-test gaps across datasets, indicating limited overfitting under few-shot conditions.

\begin{figure}[t]
\centering
\includegraphics[width=0.95\linewidth]{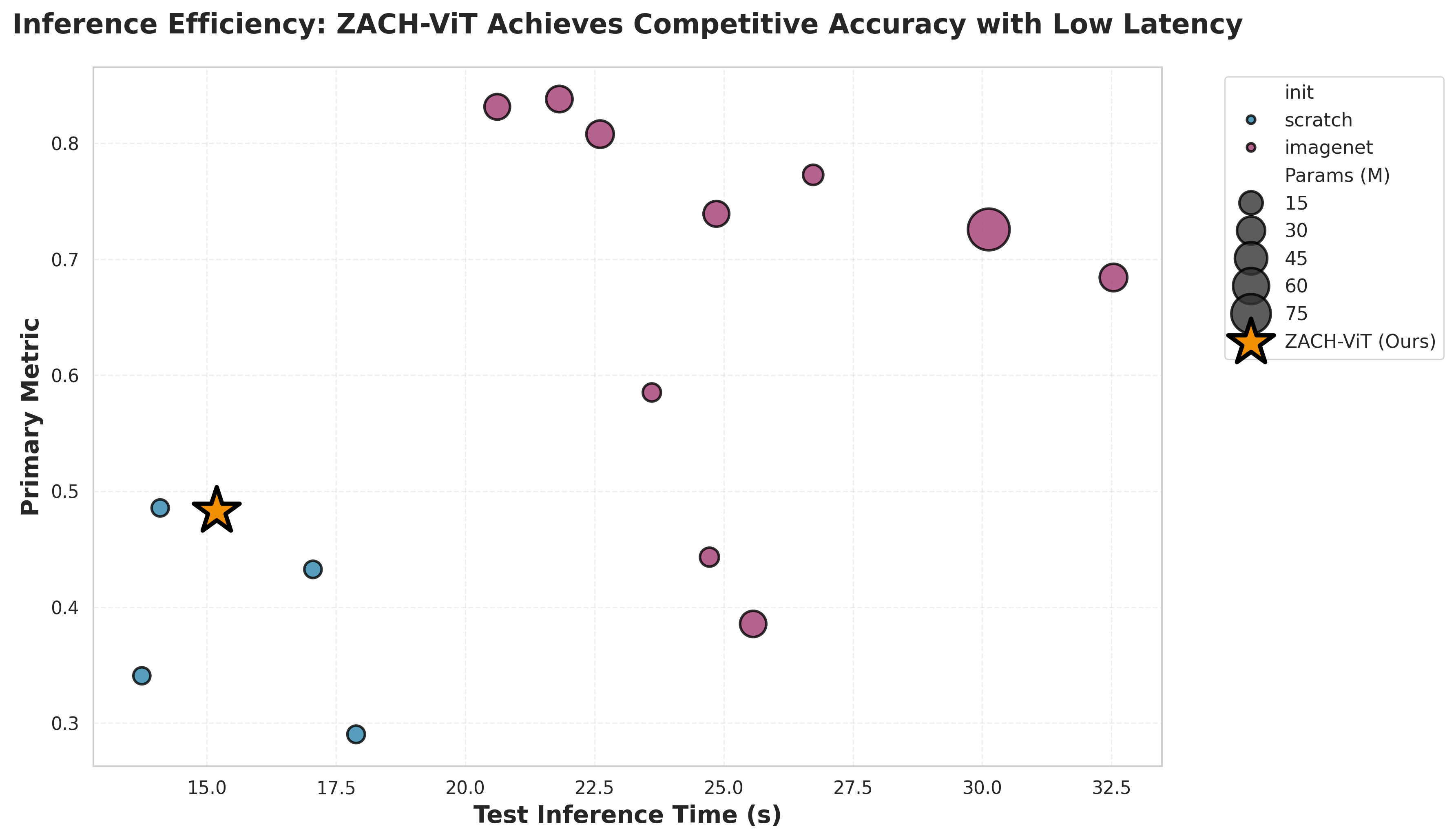}
\caption{Inference time versus performance. \model\ occupies an efficient region of the accuracy-latency trade-off.}
\label{fig:inference_efficiency}
\end{figure}

Figure~\ref{fig:inference_efficiency} further shows that \model\ occupies a favorable region of the performance--efficiency trade-off relative to substantially larger transformer models. Combined with its 0.25M parameter count and 2.95MB serialized footprint (Table~\ref{tab:all_models}), this highlights its suitability for compact medical vision settings.

\subsection{Global Ranking Analysis and Statistical Comparison}
\label{subsec:ranking_analysis}

To obtain a dataset-agnostic comparison, models were ranked per dataset using AUC@0.5 (binary tasks) or MacroF1 (multi-class tasks). Mean ranks were then computed across all seven datasets.

Statistical differences were assessed using the Friedman test followed by the Nemenyi post-hoc procedure~\cite{demsar2006statistical}. The Friedman test rejected the null hypothesis of equal performance ($p < 0.05$), motivating post-hoc comparison.

\begin{figure}[t]
\centering
\includegraphics[width=\linewidth]{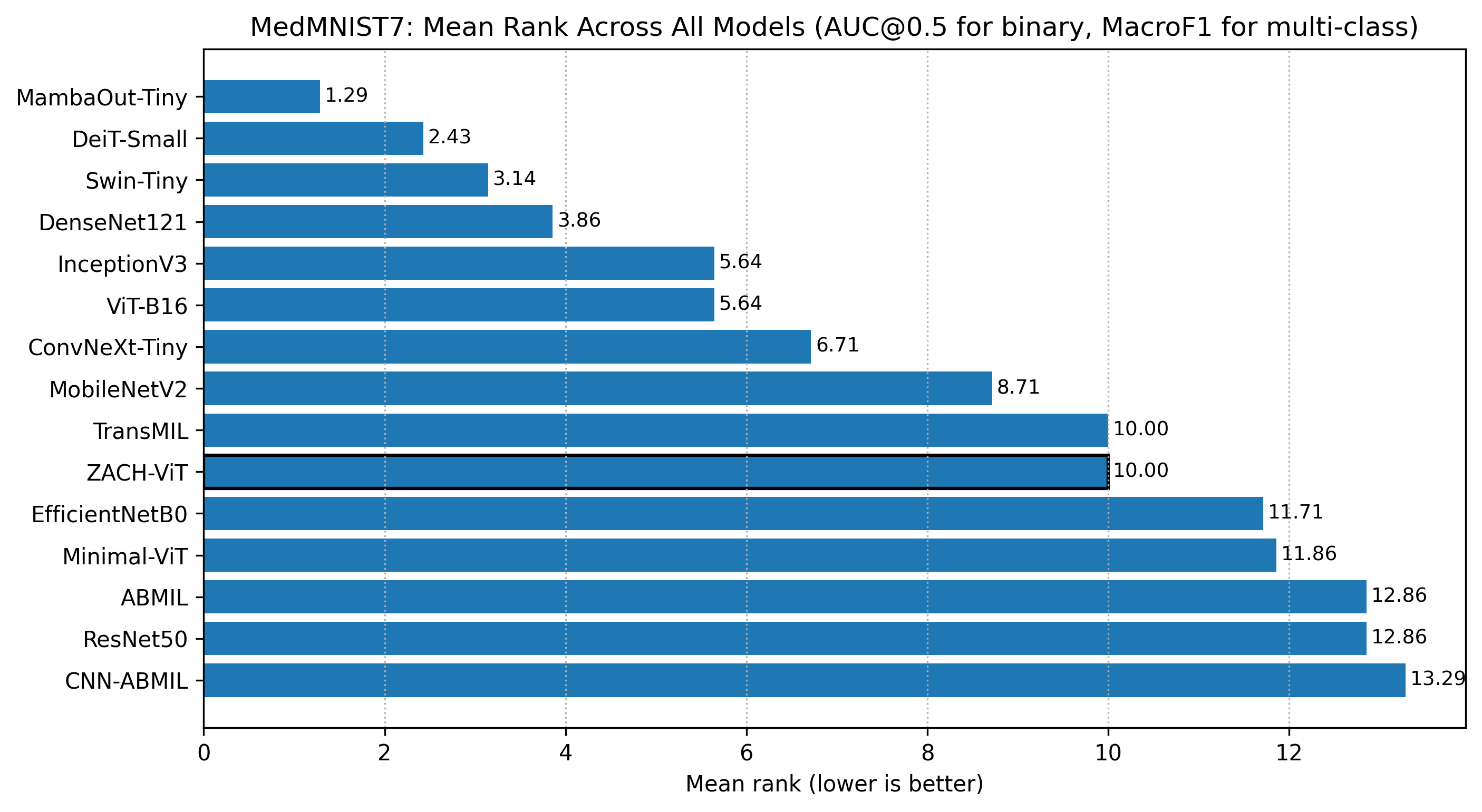}
\caption{Mean rank across all fifteen models over seven MedMNIST datasets. Lower rank indicates better average performance.}
\label{fig:mean_rank_all}
\end{figure}

Figure~\ref{fig:mean_rank_all} shows mean ranks across all models. Pretrained architectures dominate aggregate ranking, with MambaOut-Tiny achieving the best mean rank (1.29), followed by DeiT-Small (2.43) and Swin-Tiny (3.14). DenseNet121 was the strongest CNN baseline (3.86). ZACH-ViT and TransMIL obtain identical mean ranks (10.00), indicating comparable aggregate performance across heterogeneous datasets. Despite their similar aggregate ranks, the two architectures differ substantially in design philosophy: TransMIL acts as a transformer-based MIL aggregator, whereas ZACH-ViT is a compact end-to-end backbone explicitly optimized for permutation-invariant few-shot learning under strict parameter constraints.

\begin{figure}[t]
\centering
\includegraphics[width=\linewidth]{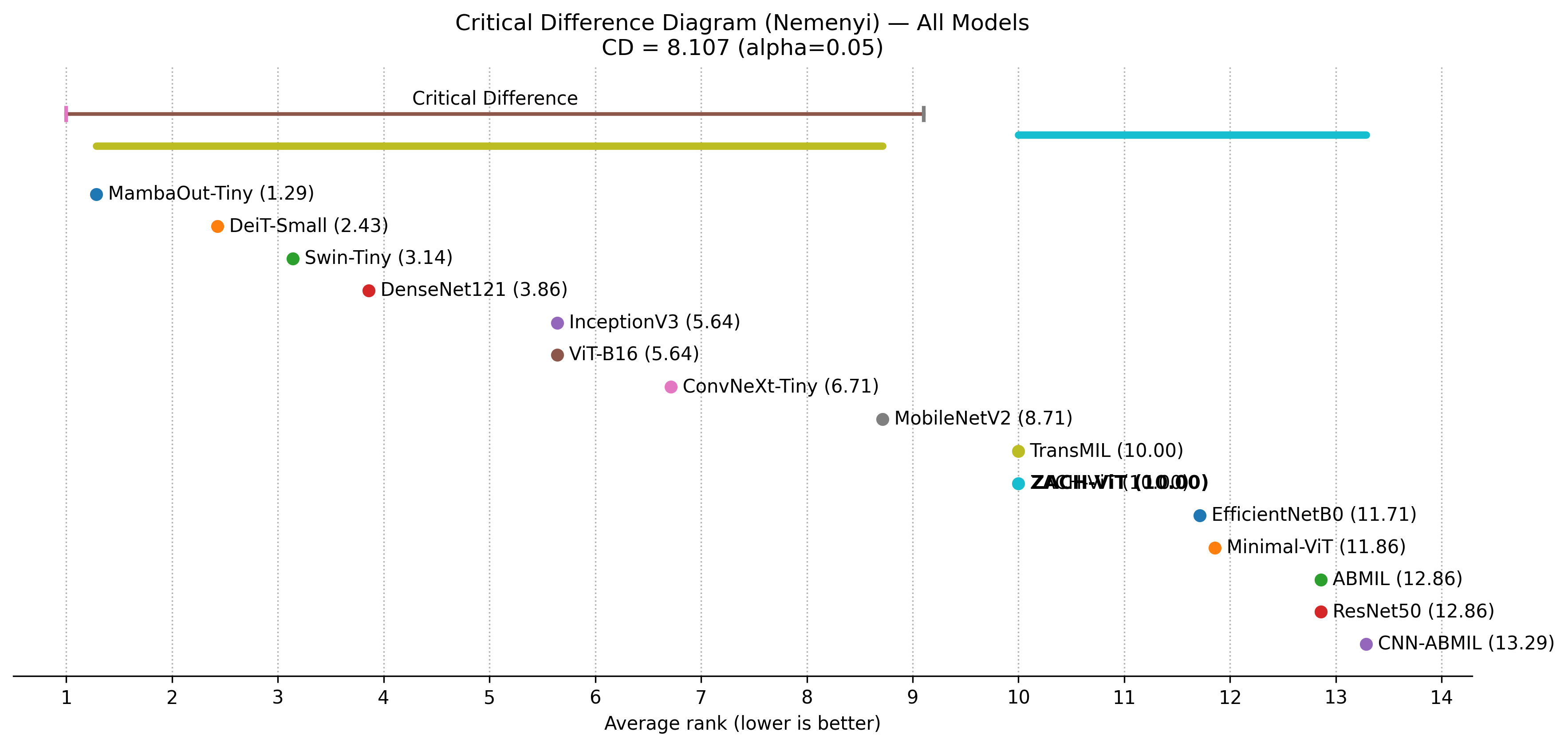}
\caption{Critical Difference diagram (Nemenyi test, $\alpha=0.05$) for all models. Connected groups indicate non-significant differences.}
\label{fig:cd_all}
\end{figure}

The corresponding Critical Difference diagram (Figure~\ref{fig:cd_all}) shows that only large rank differences reach significance due to the relatively large number of models ($k=15$) compared against seven datasets.

\begin{figure}[t]
\centering
\includegraphics[width=\linewidth]{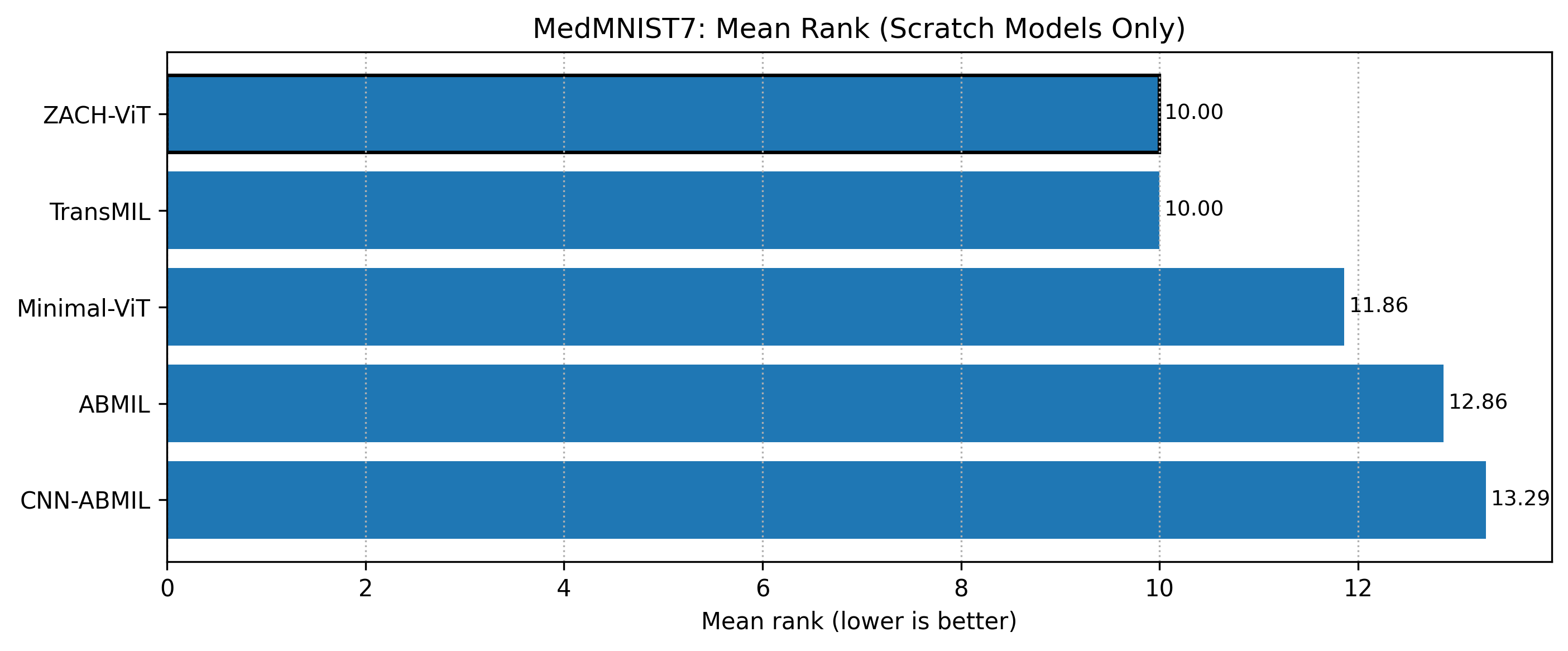}
\caption{Mean rank restricted to scratch-trained models.}
\label{fig:mean_rank_scratch}
\end{figure}

\begin{figure}[t]
\centering
\includegraphics[width=\linewidth]{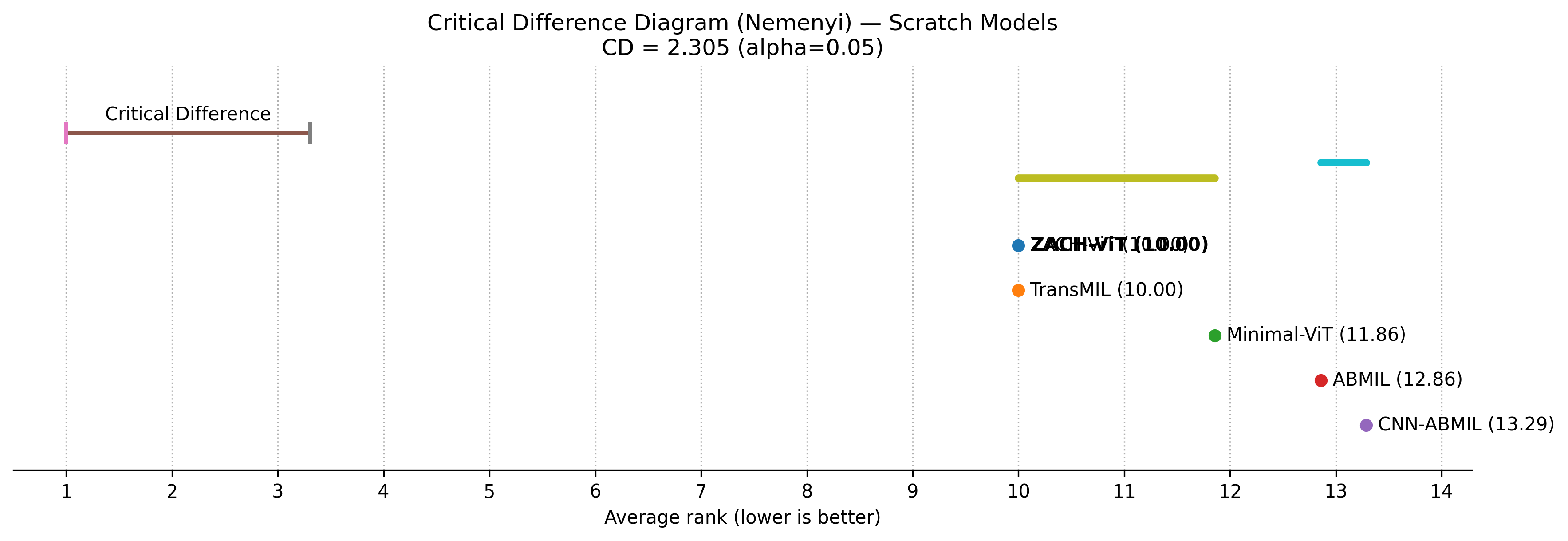}
\caption{Critical Difference diagram (Nemenyi test, $\alpha=0.05$) for scratch-trained models.}
\label{fig:cd_scratch}
\end{figure}

To isolate inductive-bias effects independent of pretraining, rankings were recomputed for scratch-trained models only (Figure~\ref{fig:mean_rank_scratch}). TransMIL and ZACH-ViT form the top group, outperforming Minimal-ViT, ABMIL, and CNN-ABMIL. The associated CD diagram (Figure~\ref{fig:cd_scratch}) indicates that TransMIL and ZACH-ViT are statistically indistinguishable under this setting.

These ranking results complement the regime-spectrum analysis: pretrained models dominate aggregate rankings, whereas permutation-invariant architectures show clear advantages in weak spatial-structure regimes.

\subsection{Limitations and Honest Assessment}
\label{subsec:limitations}

\model\ does not universally outperform all baselines. On datasets with strong anatomical structure (e.g., OCTMNIST and OrganAMNIST), models retaining positional priors or stronger pretrained representations can achieve higher performance. This behavior is consistent with the trade-off introduced by removing explicit spatial encoding: the same inductive bias that is advantageous when spatial layout is weakly informative may become restrictive when anatomical organization is diagnostically important.

A second limitation is scope. Although the evaluation spans seven medical imaging datasets and a broad range of architectures, all experiments are conducted within the MedMNIST ecosystem and under a controlled few-shot protocol. The present study therefore establishes a regime-dependent architectural trend rather than a universal statement about all medical imaging tasks or all transformer backbones. Additional validation on larger clinical datasets, especially beyond MedMNIST-style benchmark curation, would further clarify the generality of the conclusions.

Finally, the work should not be interpreted as arguing that positional priors are unnecessary in vision transformers. Rather, the evidence suggests that their usefulness depends on the structural properties of the target data. In this sense, \model\ is best viewed not as a universally superior compact transformer, but as an architecture that makes explicit the importance of matching inductive bias to spatial-structure regime.

\section{Ablation and Sensitivity Analysis}
\label{sec:ablation_sensitivity}
To assess whether the observed performance differences arise from principled architectural design rather than favorable hyperparameter choices, we complement the main benchmarking results with a three-part analysis. First, we perform a hyperparameter sensitivity study on BloodMNIST, corresponding to the weakest spatial-structure regime in our spectrum. Second, we introduce a \textbf{component ablation study} across BloodMNIST, PathMNIST, and OCTMNIST to isolate the contribution of positional embeddings, class-token aggregation, adaptive residual projections, and patch-order perturbation. Third, we conduct a \textbf{pooling operator ablation} on the same dataset subset to evaluate whether global average pooling is the most appropriate aggregation mechanism for the proposed zero-token design.

This progression moves from local sensitivity analysis in the weakest-structure regime to cross-regime architectural validation. Together, these experiments test not only whether ZACH-ViT is competitive, but also \emph{why} it works and \emph{under which structural conditions} its design choices remain advantageous.

\subsection{Hyperparameter Sensitivity in the Weakest Spatial-Structure Regime}
\label{sec:hparam_sensitivity}

To assess whether performance differences arise from architectural design rather than favorable hyperparameter choices, we conduct a sensitivity analysis across twelve architectural variants on BloodMNIST, representing the weakest spatial-structure regime in our spectrum. All variants follow the same training protocol (50 samples per class, five seeds $\{3,5,7,11,13\}$, batch size 16, 23 epochs), while varying only patch size, number of attention heads, transformer unit depth/width, and MLP configuration (Table~\ref{tab:hparam_sensitivity}). This setup isolates architectural effects while controlling for initialization variability.

\begin{table}[t]
\centering
\caption{Hyperparameter sensitivity analysis on BloodMNIST. Test MacroF1 and accuracy are reported as mean $\pm$ std across five seeds. Bold indicates statistically significant after correction for multiple comparisons over the baseline (two-tailed paired t-test, $p<0.01$, Bonferroni correction applied across all 11 variant-vs-baseline comparisons). The baseline configuration (PS=16, H=8, TU=[128,64]) represents the primary parameter-efficient setting at 0.25M parameters. All values correspond to the final experimental protocol (23 epochs, five seeds) and replace preliminary exploratory runs used during early development.}
\label{tab:hparam_sensitivity}
\resizebox{\textwidth}{!}{
\begin{tabular}{lccccccc}
\toprule
\textbf{Variant} & \textbf{Patch} & \textbf{Heads} & \textbf{TU} & \textbf{MLP} & \textbf{Params (M)} & \textbf{Test MacroF1} & \textbf{Test Acc} \\
\midrule
Baseline & 16 & 8 & 128-64 & 128-64 & 0.25 & 0.561 $\pm$ 0.051 & 0.609 $\pm$ 0.054 \\
\textbf{PS=8} & \textbf{8} & \textbf{8} & \textbf{128-64} & \textbf{128-64} & \textbf{0.17} & \textbf{0.609 $\pm$ 0.038} & \textbf{0.640 $\pm$ 0.035} \\
PS=32 & 32 & 8 & 128-64 & 128-64 & 0.54 & 0.457 $\pm$ 0.034 & 0.450 $\pm$ 0.037 \\
H=4 & 16 & 4 & 128-64 & 128-64 & 0.25 & 0.579 $\pm$ 0.039 & 0.609 $\pm$ 0.041 \\
Deeper TU & 16 & 8 & 128-128-64 & 128-64 & 0.33 & 0.497 $\pm$ 0.095 & 0.516 $\pm$ 0.102 \\
Wider TU & 16 & 8 & 256-128 & 128-64 & 0.75 & 0.638 $\pm$ 0.036 & 0.673 $\pm$ 0.038 \\
Wider MLP & 16 & 8 & 128-64 & 256-128 & 0.28 & 0.599 $\pm$ 0.036 & 0.626 $\pm$ 0.038 \\
PS=8 + H=4 & 8 & 4 & 128-64 & 128-64 & 0.17 & 0.599 $\pm$ 0.031 & 0.625 $\pm$ 0.033 \\
PS=32 + H=4 & 32 & 4 & 128-64 & 128-64 & 0.54 & 0.435 $\pm$ 0.026 & 0.438 $\pm$ 0.028 \\
Deeper+Wider & 16 & 8 & 128-128-64 & 256-128 & 0.37 & 0.552 $\pm$ 0.036 & 0.575 $\pm$ 0.038 \\
\textbf{PS=8 + Wider TU} & \textbf{8} & \textbf{8} & \textbf{256-128} & \textbf{128-64} & \textbf{0.60} & \textbf{0.684 $\pm$ 0.088} & \textbf{0.713 $\pm$ 0.092} \\
Wider TU + H=4 & 16 & 4 & 256-128 & 128-64 & 0.75 & 0.630 $\pm$ 0.048 & 0.662 $\pm$ 0.050 \\
\bottomrule
\end{tabular}
}
\end{table}

\begin{figure}[t]
\centering
\includegraphics[width=0.95\textwidth]{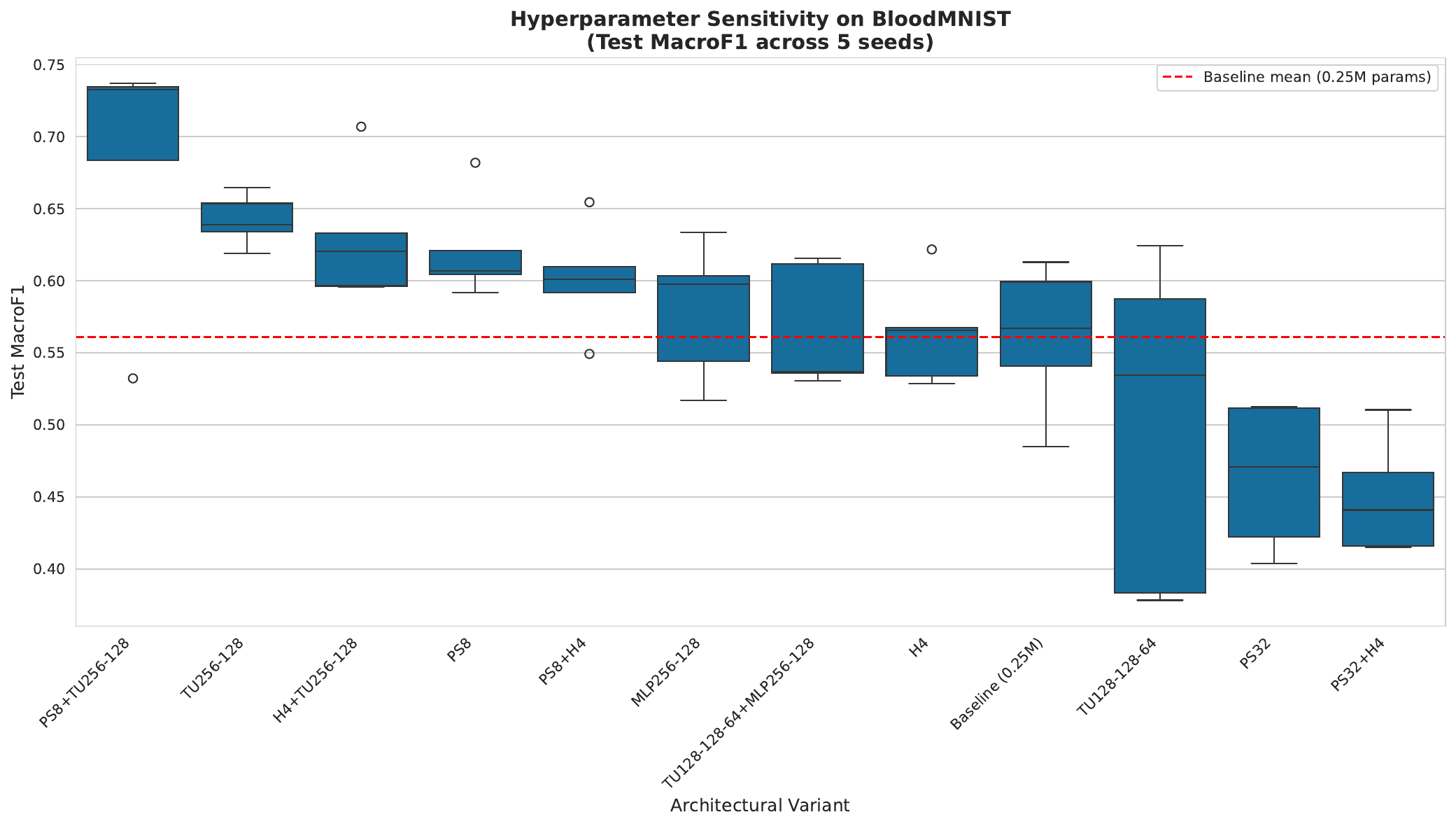}
\caption{Hyperparameter sensitivity on BloodMNIST showing test MacroF1 distributions across five random seeds. Smaller patch sizes generally yield higher performance.}
\label{fig:hparam_boxplot}
\end{figure}

\begin{figure}[t]
\centering
\includegraphics[width=0.75\textwidth]{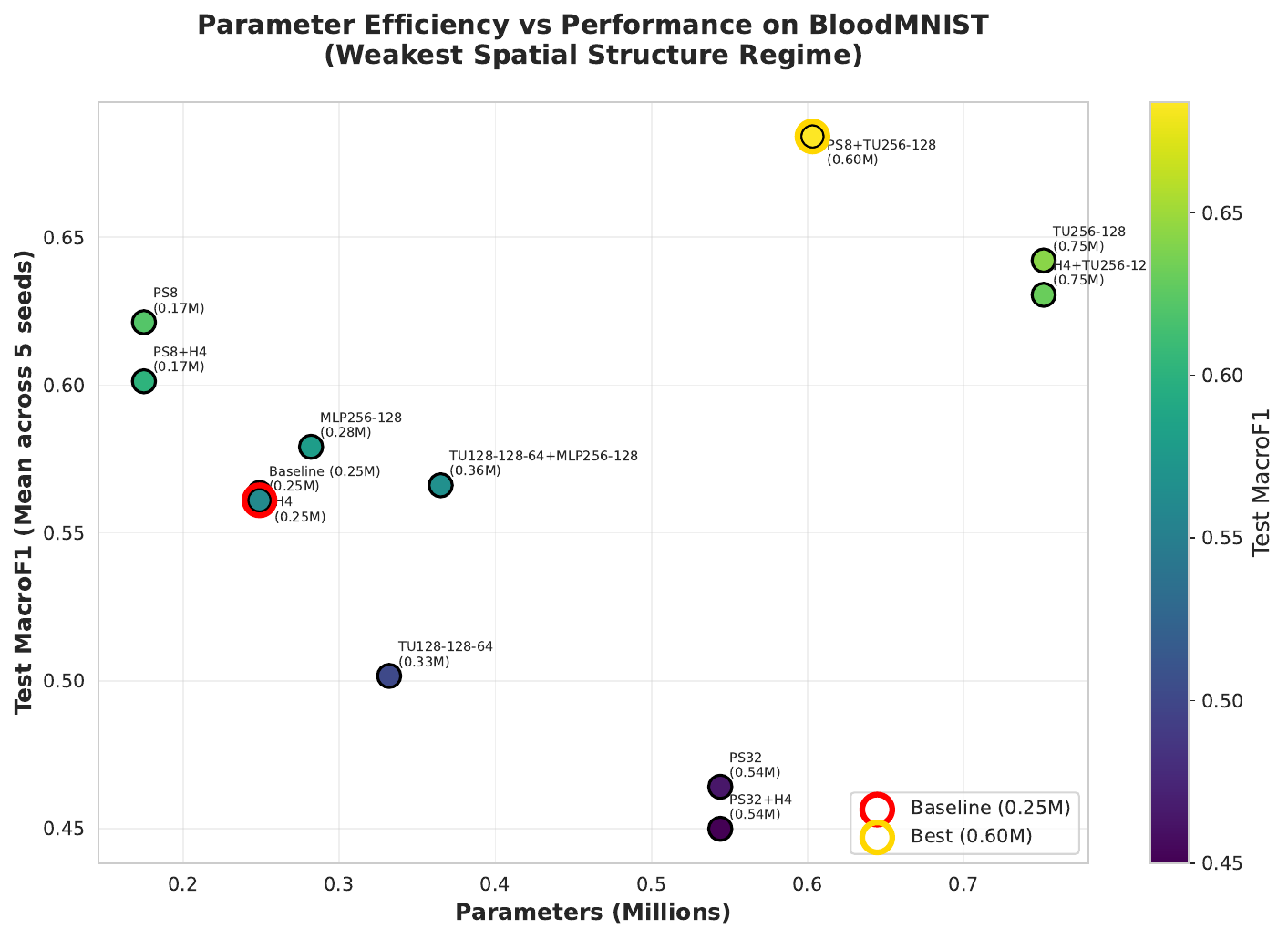}
\caption{Parameter efficiency versus performance on BloodMNIST. Higher-performing configurations often require increased capacity, while the baseline provides a favorable efficiency-performance trade-off.}
\label{fig:param_efficiency_scatter}
\end{figure}

Three observations emerge from this analysis (Figures~\ref{fig:hparam_boxplot}-\ref{fig:param_efficiency_scatter}):

\textbf{Patch size strongly influences performance.}
Reducing patch size from $16\times16$ to $8\times8$ improves MacroF1 (+0.048 over baseline, $p=0.008$), whereas increasing patch size to $32\times32$ decreases performance ($-0.104$, $p<0.001$). This trend suggests that fine-grained patch decomposition better preserves discriminative local features in weak-structure data such as blood microscopy. Performance improvements are consistent across seeds (Figure~\ref{fig:hparam_boxplot}), indicating robustness to initialization.

\textbf{Capacity scaling improves absolute performance but reduces efficiency.}
The PS=8 + wider transformer unit configuration achieves the highest MacroF1 ($0.684 \pm 0.088$) but requires substantially more parameters (0.60M vs.\ 0.25M baseline). In contrast, the PS=8 configuration with reduced parameter count (0.17M) already improves performance relative to baseline, suggesting that architectural alignment contributes more strongly than raw capacity. Figure~\ref{fig:param_efficiency_scatter} illustrates this trade-off between accuracy and parameter efficiency.

\textbf{Training remains stable across configurations.}
Most variants show low variance across seeds, indicating stable optimization. Increased depth (TU=[128,128,64]) leads to higher variability without consistent performance gains, suggesting limited benefit from additional hierarchy under weak spatial structure.

Overall, this local sensitivity analysis indicates that performance differences are primarily associated with architectural choices—particularly patch granularity—rather than random initialization or isolated hyperparameter effects. The results further support the regime-dependent interpretation observed in the main experiments: permutation-invariant modeling benefits from configurations that preserve fine-grained local information when spatial layout is weakly informative.

\subsection{Component Ablation Across Weak-to-Strong Spatial Structure}
\label{sec:component_ablation}

The preceding sensitivity analysis evaluates the stability of the selected training regime, but it does not directly isolate which architectural components are responsible for the observed behavior. We therefore perform a targeted component ablation study across three datasets spanning the structural spectrum considered in this work: BloodMNIST (weak spatial structure), PathMNIST (intermediate structure), and OCTMNIST (stronger anatomical organization). All variants are trained under the same few-shot protocol as the main experiments (50 samples per class, batch size 16, 23 epochs, five seeds $\{3,5,7,11,13\}$), while varying only a single architectural component at a time.

The evaluated variants are: (i) the full ZACH-ViT model, (ii) a version with learnable positional embeddings added after patch projection, (iii) a version using a dedicated \texttt{[CLS]} token for sequence aggregation, (iv) a version without adaptive residual projections, and (v) a random patch-shuffling probe in which patch order is permuted independently within each image. The shuffling condition serves as a direct permutation test: if performance remains stable under patch-order randomization, the task is likely not relying strongly on global spatial arrangement.

\begin{table*}[t]
\centering
\caption{Component ablation study across BloodMNIST, PathMNIST, and OCTMNIST. Values are reported as mean $\pm$ std across five seeds. Bold indicates the best mean Test MacroF1 within each dataset; for Test Accuracy, bold indicates the best mean value within each dataset.}
\label{tab:component_ablation}
\resizebox{\textwidth}{!}{
\begin{tabular}{llcc}
\toprule
\textbf{Dataset} & \textbf{Variant} & \textbf{Test MacroF1} & \textbf{Test Acc} \\
\midrule
BloodMNIST & Full ZACH-ViT & 0.605 $\pm$ 0.029 & 0.630 $\pm$ 0.041 \\
 & + Positional & 0.568 $\pm$ 0.073 & 0.592 $\pm$ 0.083 \\
 & - Adaptive Residuals & \textbf{0.612 $\pm$ 0.060} & \textbf{0.644 $\pm$ 0.075} \\
 & Random Shuffle & 0.573 $\pm$ 0.055 & 0.594 $\pm$ 0.063 \\
 & \texttt{[CLS]} token & 0.512 $\pm$ 0.088 & 0.556 $\pm$ 0.096 \\
\midrule
PathMNIST & Full ZACH-ViT & 0.554 $\pm$ 0.026 & 0.633 $\pm$ 0.030 \\
 & + Positional & \textbf{0.565 $\pm$ 0.038} & \textbf{0.650 $\pm$ 0.049} \\
 & - Adaptive Residuals & 0.550 $\pm$ 0.009 & 0.630 $\pm$ 0.009 \\
 & Random Shuffle & 0.527 $\pm$ 0.016 & 0.605 $\pm$ 0.021 \\
 & \texttt{[CLS]} token & 0.356 $\pm$ 0.033 & 0.424 $\pm$ 0.049 \\
\midrule
OCTMNIST & Full ZACH-ViT & 0.245 $\pm$ 0.015 & \textbf{0.310 $\pm$ 0.017} \\
 & + Positional & \textbf{0.250 $\pm$ 0.021} & 0.308 $\pm$ 0.020 \\
 & - Adaptive Residuals & 0.223 $\pm$ 0.030 & 0.276 $\pm$ 0.035 \\
 & Random Shuffle & 0.246 $\pm$ 0.018 & 0.302 $\pm$ 0.025 \\
 & \texttt{[CLS]} token & 0.129 $\pm$ 0.008 & 0.186 $\pm$ 0.009 \\
\bottomrule
\end{tabular}
}
\end{table*}

\begin{figure*}[t]
\centering
\includegraphics[width=\textwidth]{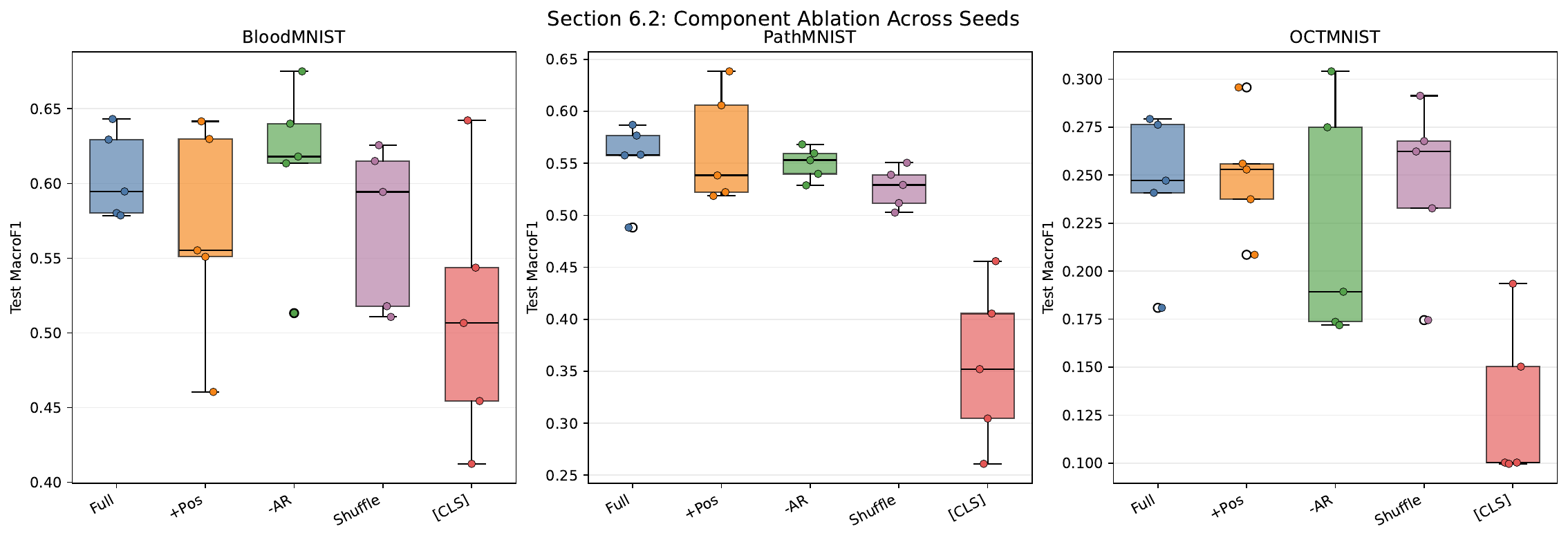}
\caption{Component ablation study across BloodMNIST, PathMNIST, and OCTMNIST showing Test MacroF1 distributions across five seeds for each architectural variant. The effect of the ablated components is regime-dependent: positional embeddings become more favorable in PathMNIST and OCTMNIST, while the \texttt{[CLS]} token is consistently associated with lower performance.}
\label{fig:component_ablation_boxplots}
\end{figure*}

\begin{figure}[t]
\centering
\includegraphics[width=0.9\columnwidth]{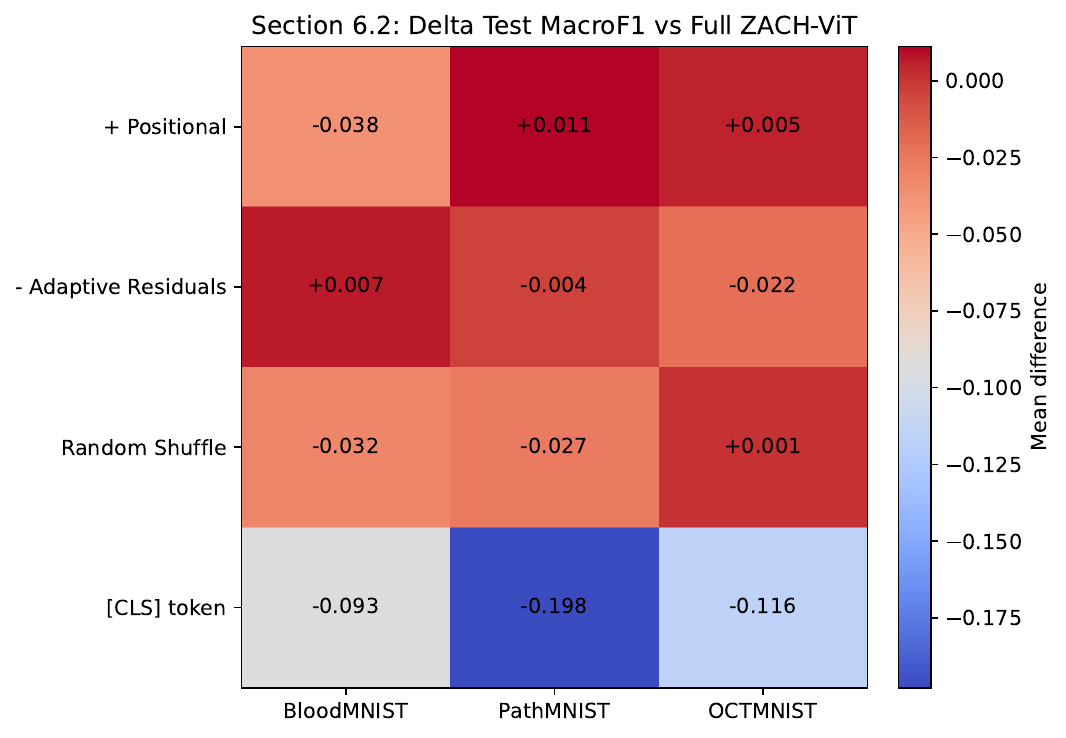}
\caption{Change in mean Test MacroF1 relative to the full ZACH-ViT for each ablation variant and dataset. Positive values indicate improvement over the full model and negative values indicate degradation. The figure highlights the mild benefit of positional embeddings in the more structured regimes and the consistently negative effect of the \texttt{[CLS]} token.}
\label{fig:component_ablation_delta}
\end{figure}

Several observations emerge from this analysis (Figures~\ref{fig:component_ablation_boxplots}-\ref{fig:component_ablation_delta}).

\textbf{The effect of architectural components is regime-dependent rather than uniform.}
On BloodMNIST, the best-performing variant is the version without adaptive residual projections, while the full ZACH-ViT remains close behind. On PathMNIST, adding positional embeddings yields the highest mean Test MacroF1 and accuracy. On OCTMNIST, positional embeddings again provide the highest mean Test MacroF1, whereas the full model achieves the highest mean accuracy. Taken together, these results support a regime-dependent interpretation: the usefulness of explicit ordering mechanisms increases as the task becomes more structurally organized.

\textbf{Positional information is least useful in the weakest-structure regime and more beneficial as structure increases.}
Adding positional embeddings does not improve over the best BloodMNIST configuration, but it becomes the strongest MacroF1 variant on both PathMNIST and OCTMNIST. This pattern is consistent with the view that explicit spatial priors are unnecessary when local appearance dominates, yet become mildly beneficial when tissue organization or anatomical layout carries more stable discriminative information.

\textbf{The \texttt{[CLS]} token is consistently unfavorable, especially outside BloodMNIST.}
Reintroducing a dedicated sequence-level aggregation token yields the lowest mean Test MacroF1 on all three datasets, with the strongest degradation on PathMNIST and OCTMNIST. This suggests that, in the compact regime studied here, restoring a conventional ordered token-aggregation mechanism is not beneficial and may interfere with the intended permutation-invariant inductive bias.

\textbf{Patch shuffling remains relatively competitive, particularly at the two ends of the structure spectrum.}
The random patch-shuffling probe remains close to the full model on BloodMNIST and OCTMNIST, and shows only moderate degradation on PathMNIST. These results indicate that a substantial fraction of the discriminative signal is still captured by local appearance statistics rather than by rigid global patch ordering. Even in the more structured regimes, performance does not appear to depend exclusively on precise token order.

Overall, the component ablation study refines the central claim of the manuscript. Rather than supporting a simplistic conclusion that permutation-invariant processing is always superior, the results indicate a more nuanced regime-dependent picture. In weakly structured data, the full ZACH-ViT design remains highly competitive without requiring explicit positional support; in more structured regimes, limited positional information becomes mildly useful; and in all regimes, reintroducing a dedicated \texttt{[CLS]} token is detrimental. This strengthens the argument that the benefit of ZACH-ViT lies not in universally removing all ordering mechanisms, but in matching the amount of architectural bias to the structural properties of the target task.

\subsection{Pooling Operator Ablation}
\label{sec:pooling_ablation}

The proposed ZACH-ViT replaces the conventional class-token aggregation used in many Vision Transformers with global average pooling over the final patch representations. This is a central design choice: it removes the need for a dedicated aggregation token and preserves symmetry with respect to patch ordering. To determine whether this choice is empirically justified, we compare four aggregation strategies: global average pooling (GAP), global max pooling, attention-based pooling, and class-token pooling. As in Section~\ref{sec:component_ablation}, evaluation is performed on BloodMNIST, PathMNIST, and OCTMNIST under the same few-shot protocol.

\begin{table*}[t]
\centering
\caption{Pooling operator ablation across BloodMNIST, PathMNIST, and OCTMNIST. Values are reported as mean $\pm$ std across five seeds. Bold indicates the best mean Test MacroF1 within each dataset; for Test Accuracy, bold indicates the best mean value within each dataset.}
\label{tab:pooling_ablation}
\resizebox{\textwidth}{!}{
\begin{tabular}{llcc}
\toprule
\textbf{Dataset} & \textbf{Pooling Operator} & \textbf{Test MacroF1} & \textbf{Test Acc} \\
\midrule
BloodMNIST & Global Average Pooling (GAP) & \textbf{0.605 $\pm$ 0.029} & \textbf{0.630 $\pm$ 0.041} \\
 & Attention Pooling & 0.584 $\pm$ 0.039 & 0.603 $\pm$ 0.057 \\
 & Global Max Pooling & 0.551 $\pm$ 0.080 & 0.566 $\pm$ 0.091 \\
 & \texttt{[CLS]} Token & 0.471 $\pm$ 0.089 & 0.497 $\pm$ 0.105 \\
\midrule
PathMNIST & Global Average Pooling (GAP) & 0.554 $\pm$ 0.039 & \textbf{0.637 $\pm$ 0.030} \\
 & Attention Pooling & \textbf{0.557 $\pm$ 0.039} & 0.634 $\pm$ 0.042 \\
 & Global Max Pooling & 0.425 $\pm$ 0.065 & 0.516 $\pm$ 0.077 \\
 & \texttt{[CLS]} Token & 0.369 $\pm$ 0.111 & 0.464 $\pm$ 0.094 \\
\midrule
OCTMNIST & Global Average Pooling (GAP) & 0.245 $\pm$ 0.040 & 0.310 $\pm$ 0.017 \\
 & Attention Pooling & \textbf{0.249 $\pm$ 0.023} & \textbf{0.315 $\pm$ 0.038} \\
 & Global Max Pooling & 0.189 $\pm$ 0.049 & 0.274 $\pm$ 0.008 \\
 & \texttt{[CLS]} Token & 0.107 $\pm$ 0.009 & 0.248 $\pm$ 0.010 \\
\bottomrule
\end{tabular}
}
\end{table*}

\begin{figure*}[t]
\centering
\includegraphics[width=\textwidth]{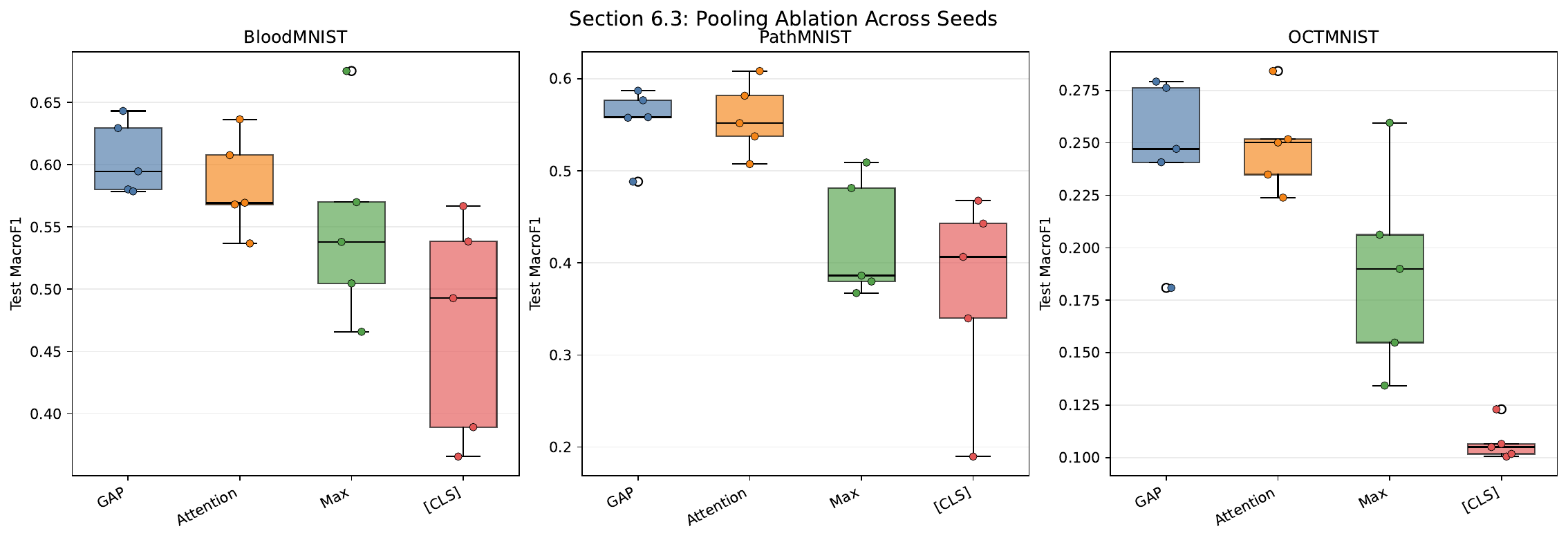}
\caption{Pooling operator ablation across BloodMNIST, PathMNIST, and OCTMNIST showing Test MacroF1 distributions across five seeds for each aggregation strategy. GAP is the strongest default overall, attention pooling becomes mildly competitive in the more structured regimes, and \texttt{[CLS]} pooling is consistently associated with the lowest performance.}
\label{fig:pooling_ablation_boxplots}
\end{figure*}

\begin{figure}[t]
\centering
\includegraphics[width=0.9\columnwidth]{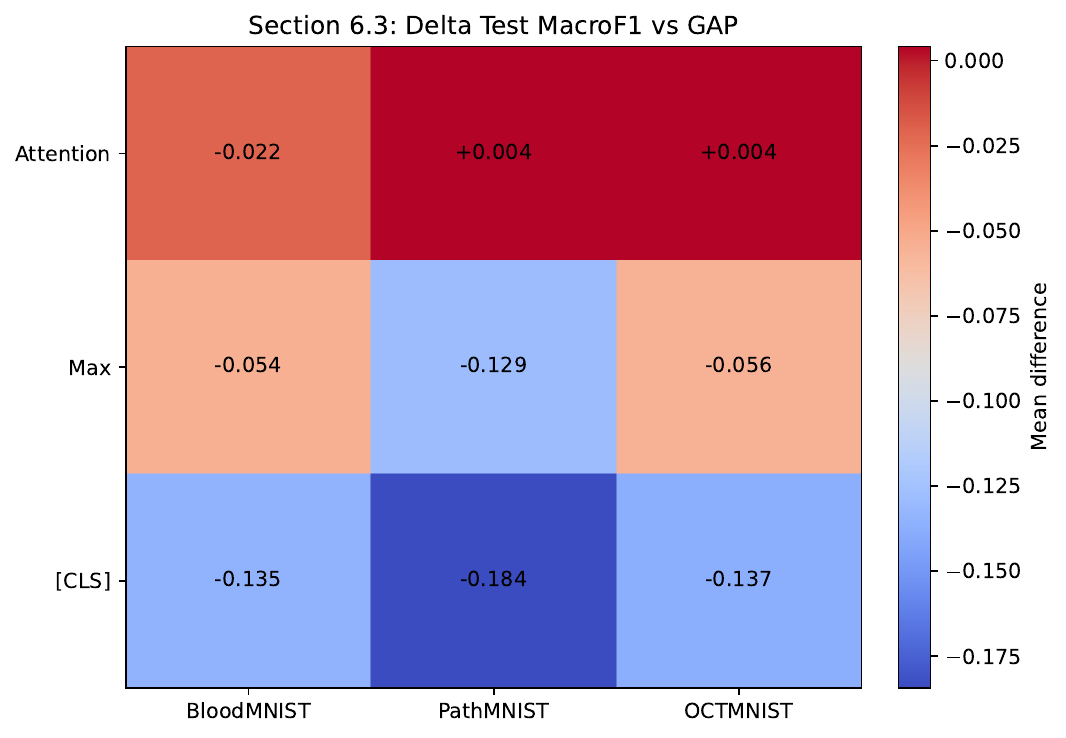}
\caption{Change in mean Test MacroF1 relative to GAP for each pooling variant and dataset. Positive values indicate improvement over GAP and negative values indicate degradation. The figure highlights the small gains of attention pooling in the more structured regimes and the consistently negative effect of max and \texttt{[CLS]} pooling.}
\label{fig:pooling_ablation_delta}
\end{figure}

Several observations emerge from this analysis (Figures~\ref{fig:pooling_ablation_boxplots}-\ref{fig:pooling_ablation_delta}).

\textbf{GAP is the strongest and most robust default aggregation strategy overall.}
On BloodMNIST, GAP achieves the best mean Test MacroF1 and accuracy, clearly outperforming max pooling and class-token aggregation. On PathMNIST, GAP remains essentially tied with attention pooling in MacroF1 while yielding slightly higher mean accuracy. On OCTMNIST, GAP again remains highly competitive, only marginally below attention pooling in MacroF1. These results support the central architectural choice of ZACH-ViT: symmetric averaging over tokens provides a strong and stable default across structurally distinct regimes.

\textbf{Attention pooling becomes mildly competitive as structure increases.}
Attention pooling is the best MacroF1 variant on PathMNIST and OCTMNIST, although the gains over GAP are small. This suggests that when structural organization becomes more informative, learned token weighting may recover a modest advantage by emphasizing particularly salient regions. At the same time, the small margin over GAP indicates that the proposed average-based aggregation remains well aligned with the underlying tasks.

\textbf{Global max pooling and class-token aggregation are consistently inferior.}
Global max pooling underperforms GAP and attention pooling on all three datasets, suggesting that hard selection over tokens overemphasizes isolated local responses at the expense of distributed evidence. Class-token pooling is the weakest option overall, with the lowest mean Test MacroF1 in every dataset and the largest degradation in the more structured regimes. This mirrors the component-ablation findings and further indicates that reintroducing a dedicated ordered summary token is not beneficial in the compact setting considered here.

\textbf{The pooling results reinforce the regime-dependent interpretation without overturning the main design choice.}
The results do not suggest that one should abandon GAP in favor of more complex aggregation. Rather, they indicate that GAP is the most robust and principled default, while attention pooling may become marginally favorable when structure increases. This is consistent with the broader thesis of the paper: compact permutation-invariant transformers are especially well suited to weakly structured data, while modest learned weighting can become useful as tasks move toward stronger anatomical organization.

Taken together, the results of Sections~\ref{sec:component_ablation} and~\ref{sec:pooling_ablation} provide a more complete picture of the proposed architecture. Section~\ref{sec:component_ablation} isolates the contribution of key design components, while Section~\ref{sec:pooling_ablation} evaluates whether the final token aggregation mechanism is aligned with the structural properties of the target task. 

\section{Discussion}
\label{sec:discussion}

Our validation across seven medical datasets, together with targeted sensitivity and ablation analyses, provides empirical support for the architectural hypothesis underlying \model. Rather than pursuing universal benchmark dominance, the results reveal a consistent relationship between architectural behavior and data structure. The regime spectrum analysis shows that \model's relative advantage is largest when spatial layout is weakly informative and decreases as spatial organization becomes more diagnostically relevant. Taken together, these findings suggest that architectural evaluation should explicitly account for spatial-structure heterogeneity rather than relying solely on aggregate benchmark performance.

\textbf{Architectural alignment can outweigh scale under data scarcity.}
While pretrained models generally outperform scratch-trained ones in low-data settings, \model\ narrows this gap substantially within its parameter regime (0.25M versus 2-85M). On BloodMNIST and PathMNIST, where spatial order is weakly informative, it even exceeds substantially larger models in parameter efficiency. This suggests that in data-scarce medical imaging scenarios, architectural parsimony aligned with data structure can matter more than brute-force scaling.

\textbf{Permutation-invariant processing is beneficial, but not universally.}
The main contribution of \model\ is not to argue that positional priors should always be removed. Rather, the results indicate that their usefulness depends on regime. The component ablation shows that positional support becomes mildly beneficial as structure increases, especially on PathMNIST and OCTMNIST, while the reintroduction of a dedicated \texttt{[CLS]} token is consistently unfavorable. The pooling ablation further shows that global average pooling is a robust default across regimes, with attention-based pooling becoming only marginally competitive in the more structured settings. Together, these findings refine the architectural claim: the value of permutation-invariant processing lies in matching inductive bias to the spatial organization of the data.

\textbf{Generalization behavior is consistent with reduced reliance on unstable spatial priors.}
The small train-test gaps observed across datasets are consistent with the idea that removing explicit positional assumptions can reduce overfitting under few-shot conditions. When spatial layout is weakly informative, strong positional bias may encourage shortcut learning based on unstable acquisition-dependent correlations. In contrast, permutation-invariant processing encourages the model to rely more heavily on local visual evidence and compositional statistics. While this interpretation is supported by the current experiments, broader validation on larger clinical datasets will be necessary to determine how consistently this generalization pattern holds beyond MedMNIST-style benchmarks.

\textbf{Implications for compact medical vision models.}
Most ViT variants assume that spatial structure is always diagnostically useful. Our results suggest that this assumption should be treated as data-dependent rather than universal, particularly in compact models designed for low-data or resource-constrained settings. In this sense, \model\ should be interpreted less as a lightweight replacement for all ViTs and more as an example of architecture design driven by inductive-bias alignment. The broader lesson is that compactness alone is insufficient: what matters is whether model bias matches the structure of the target task.

\textbf{Remaining limitations.}
Despite the encouraging findings, the present work is still confined to MedMNIST-style benchmarks under a controlled few-shot setting. The paper therefore establishes a strong empirical trend rather than a final general theory of permutation-invariant vision transformers. Extending this analysis to larger clinical datasets, especially beyond curated benchmark suites and toward raw clinical-scale imaging data, will be important for understanding how broadly the identified regime-dependent behavior transfers.

\section{Conclusion}
\label{sec:conclusion}

We introduced \model, a zero-token Vision Transformer that removes positional embeddings and the class token, relying instead on permutation-invariant patch processing and global average pooling. Evaluation across seven representative MedMNIST datasets shows that \model\ achieves competitive performance with minimal parameters (0.25M) and no pretraining, while remaining stable under few-shot conditions where larger models often degrade. Importantly, its advantage is regime-dependent: strongest when spatial order is weakly informative (BloodMNIST, PathMNIST) and reduced when anatomical structure imposes fixed layouts (OCTMNIST, OrganAMNIST).

The contribution of this work is therefore not only a compact architecture, but also an empirical argument about inductive-bias alignment. The additional component and pooling ablations show that positional support becomes mildly useful as structure increases, whereas reintroducing a dedicated \texttt{[CLS]} token is consistently detrimental and global average pooling remains the most robust aggregation strategy overall. These findings support the view that architectural priors should be treated as data-dependent design choices rather than default components.

For medical imaging scenarios characterized by limited data, constrained compute, or weakly informative spatial layout, \model\ provides a principled alternative to conventional ViTs. More broadly, the results suggest that compact transformer design should be guided less by benchmark convention and more by the structural properties of the target domain.

\bibliographystyle{plain}

\end{document}